\documentclass[letterpaper, 10 pt, conference]{ieeeconf}
\IEEEoverridecommandlockouts                              
\usepackage[nospace,noadjust]{cite}
\usepackage{amsmath,amssymb,amsfonts}
\usepackage{algorithmic}
\usepackage{graphicx, multirow}
\usepackage{textcomp}
\usepackage{xcolor}
\usepackage{listings}
\usepackage{hyperref}
\usepackage{float}
\usepackage{todonotes}
\usepackage{booktabs}
\usepackage[ruled, lined, linesnumbered, commentsnumbered, longend]{algorithm2e}
\def\BibTeX{{\rm B\kern-.05em{\sc i\kern-.025em b}\kern-.08em
		T\kern-.1667em\lower.7ex\hbox{E}\kern-.125emX}}

\DeclareMathOperator{\notype}{\lambda}
\DeclareMathOperator{\abox}{\mathcal{A}}

\newfloat{lstfloat}{htbp}{lop}
\floatname{lstfloat}{Listing}
\title{\LARGE \bf
	Translating Universal Scene Descriptions into\\ Knowledge Graphs for Robotic Environment
}

\author{Giang Hoang Nguyen$^{1}$, Daniel Be{\ss}ler$^{1}$, Simon Stelter$^{1}$, Mihai Pomarlan$^{1}$ and Michael Beetz$^{1}$
	\thanks{$^{1}$Institute for Artificial Intelligence,
		University of Bremen, Germany}%
}

\begin{document}
	
	%
	
	\maketitle
	
	\begin{abstract}
		Robots performing human-scale manipulation tasks require an extensive amount of knowledge about their surroundings in order to perform their actions competently and human-like.
		In this work, we investigate the use of virtual reality technology as an implementation for robot environment modeling, and present a technique for translating scene graphs into knowledge bases. To this end, we take advantage of the Universal Scene Description (USD) format which is an emerging standard for the authoring, visualization and simulation of complex environments. We investigate the conversion of USD-based environment models into Knowledge Graph (KG) representations that facilitate semantic querying and integration with additional knowledge sources. 
		The contributions of the paper are validated through an application scenario in the service robotics domain.
	\end{abstract}
	
	
	\section{Introduction}
	The vision of a world with accurate digital twins for every environment holds promise for advancing the field of service robotics. With access to detailed environment models, robots would be able to interact with their surroundings in a more sophisticated manner. Advanced game engines with physics simulation capabilities are a significant step towards achieving the vision of highly detailed environment models. These models use scene graph data structures and computational techniques for real-time maintenance.
	This work aims to explore the assumption that environment models used in modern games and the associated infrastructure is appropriate for robot environment modeling.
	
	Scene graphs enable the representation of the compositional structure of environments, where each component and its configurations are modeled as sub-graphs within the scene graph. This allows selective consideration of only relevant parts of the graph based on the agent's viewpoint. 
	A robot can start with a coarse object model, refine it with additional knowledge, and insert it at different positions in the graph as the object moves. 
	Nonetheless, the use of detailed environment models presents challenges in data management, such as slower performance and increased memory usage. These challenges must be addressed in order to fully realize the potential benefits of detailed environment models in robotics.
	
	\begin{figure}[t]
		\centering
		\includegraphics[width=\columnwidth]{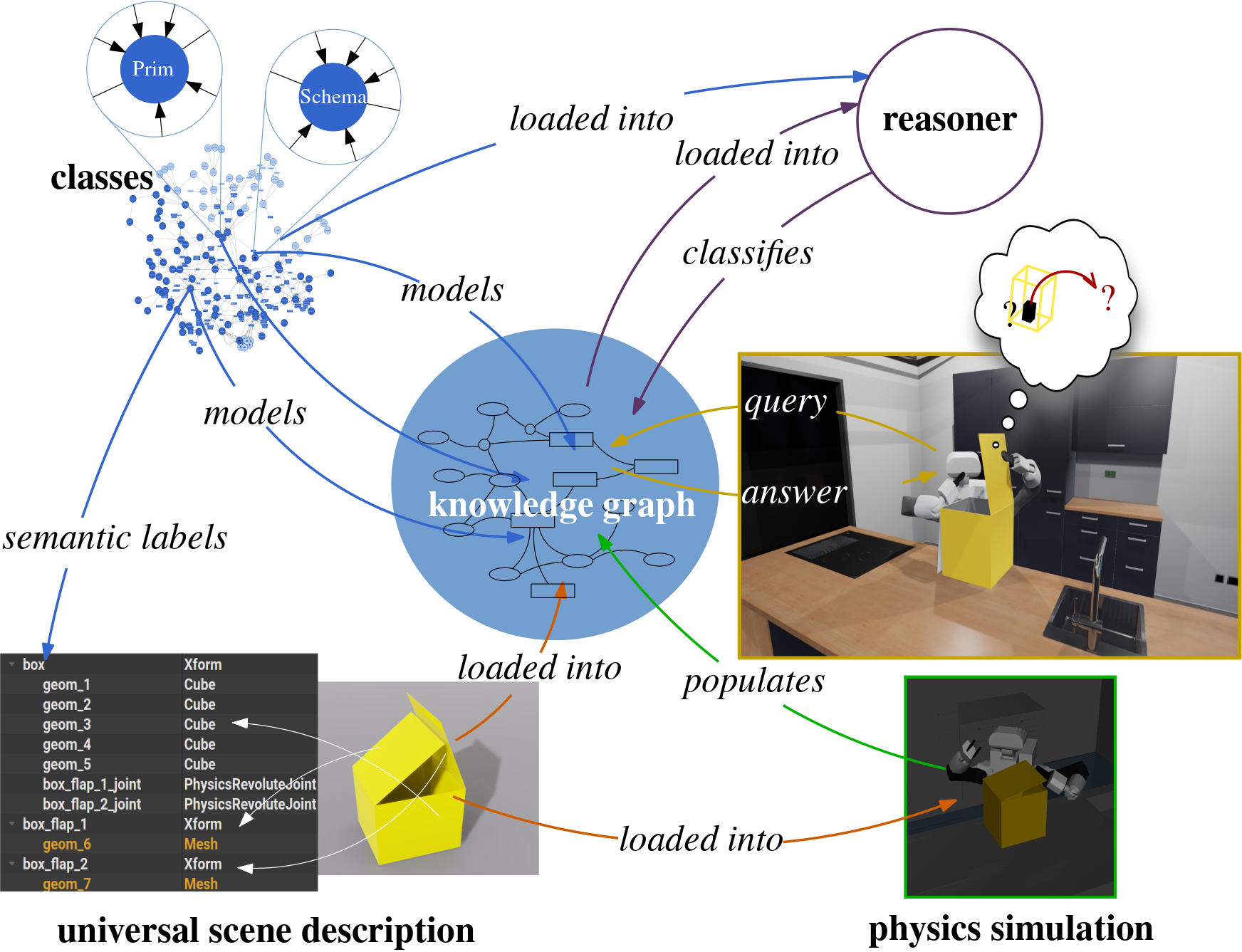}
		\caption{Interactions of a Knowledge Graph with different components: Classes, Scene Description, Physics Simulation and Reasoner}
		\label{fig:abstract}
	\end{figure}

	The \textit{Universal Scene Description} (USD) format 
	is an attempt to address these challenges \cite{blevins2018zero}.
	USD is an open-source, industry-standard format for describing and exchanging 3D scenes and 
	assets. 
	It enables artists and developers to work with complex scenes and
	assets across multiple applications and workflows.
	It further supports the simulation of scenes,
	allowing artists and developers to test and refine their virtual environments by adjusting elements such as lighting and physics without the need to export or import data between different software applications. This streamlined workflow saves time and improves the quality of the final product.
	
	To be practically useful for a robot, environment models need to be connected to the task of the robot and its body.
	Such a link is crucial to determine which objects can be used to perform a task, and how they can be handled.
	Our approach is to represent such information as a linked KG.
	The KG is composed of knowledge covering robot, environment and task.
	Here, we focus on extracting environment knowledge from USD files to provide additional knowledge required for task execution.
	
	Accordingly, the main research question motivating this work is whether standardized scene graphs represented in the USD format can be transformed into a representation that enables robots to model their environment while also rendering the model interpretable through semantic queries.
	Thus, a mapping from the USD format to a KG representation is proposed, which is fixed by an ontology that also supports the translation process via automated reasoning.
	Definitions in the ontology are further aligned with a common ontology, enabling linking to other knowledge sources. 
	The resulting KG incorporates data mapped from the USD file, inferences made by the reasoner, and dynamical data computed by a simulator in order to build a more comprehensive relational model of the environment.
	This process is depicted in Figure~\ref{fig:abstract}.
	The resulting infrastructure can be easily linked to other information sources, and provides a lowcost way of representing, manipulating and testing environments. Summarizing, the contributions of this work are the following:
	
	
	
	\begin{itemize}
		\item a formalization of the USD format with extensions for semantic tagging of its entities,
		\item an ontological model of USD that fixes a KG and supports the translation process via inference, and
		\item the design and implementation of a method for automated KG construction from USD files.
	\end{itemize}
	The contributions will be validated experimentally by constructing a KG from a USD file, using a reasoner to finalize the KG and executing a robotics task in a physics simulation to update the KG.
	
	\section{Related Work}
	
	Developing robotic agents that can perform various indoor tasks is a complex and challenging problem that requires a comprehensive understanding of the environment.
	In this section, we review the related work that emphasizes the importance of using the USD in generating 3D objects and semantic models for robotics reasoning.
	
	Mapping 3D objects and scene graphs using sensor data is an approach to understanding the physical properties of the environment. Rusu et al. \cite{rusu2009model} and Antoni et al. \cite{rosinol20203d} presented systems for acquiring accurate 3D object maps using depth cameras and inertial measurement units. Beetz et al. \cite{Beetz2022} focused on mapping objects in retail environments. However, these models often suffer from limited transferability, making it challenging to reuse them in different contexts. Moreover, the models' representation in XML-based formats or point-clouds can be difficult to import into different simulation tools and prone to errors, as noted by Ivanou et al. \cite{ivanou2021robot} and Tola et al. \cite{tola2023understanding}. The USD adresses these challenges by describing a scene graph with capability to handle large-scale \cite{jatavallabhula2019kaolin} and complex datasets \cite{zhao2022vrkitchen2}. 
	
	
	Creating accurate 3D object maps is insufficient for intelligent robotic agents. They also need to reason about potential uses of objects in their environment. Beßler et al. \cite{bessler20affordances} proposed a conceptualization of affordances based on physical properties of objects and their potential uses. Similarly, Crespo et al. \cite{crespo2017relational} presented a semantic relational model for robots to manage and query their environment. Semantic mapping is essential for creating a representation of an environment that includes both geometric and semantic properties. 
	USD's ability to represent complex scenes, objects, and their relationships can support semantic mapping and thus improve the reasoning capabilities of robotic agents.
	
	
	Recent research has introduced the application of semantic models and scene graphs in robotics reasoning (\cite{zhu2021hierarchical}, \cite{miao2023long}, \cite{agia2022taskography}). To enable comprehensive robotics reasoning with semantic models, robust knowledge processing is essential. In this regard, Haidu et al. \cite{haidu2018knowrobsim} introduced KnowRobSIM, an extension of the KnowRob architecture (\cite{tenorth2009knowrob,beetz2018know}) that encompasses subsymbolic world modeling, action simulation based on a physics engine, and scene rendering. However, this approach relies solely on scene description from Unreal Engine and its physics. Thus, the inclusion of semantic information into the 3D data represented in USD offers a promising opportunity to advance robotics reasoning in a more comprehensive manner. 

	\section{The Universal Scene Description Format}
	\label{sec:formalization}
	The USD file format stores 3D scenes as hierarchical scene graphs (or \emph{stages}).
	In USD, data is arranged hierarchically into namespaces of \emph{prims} (primitives).
	Each prim can hold child prims, as well as \emph{attributes} and \emph{relationships}, referred to as \emph{properties}.
	Attributes have typed values that can change over time, while relationships are pointers to other objects in the hierarchy.
	All of these elements, including prims and their contents, are stored in a \emph{layer}, which is a scene description container for USD.
	This hierarchical structure allows for easy organization of the scene, and enables transformations and properties to be inherited from parent prims to their children.
	Listing~\ref{lst:example} shows an example of how a cardboard box with two flaps can be represented using a hierarchy of prims in USD.
	
	\begin{lstfloat}[h]
		\begin{lstlisting}[tabsize=2, basicstyle=\scriptsize\ttfamily, language=python, morekeywords={uniform, token, rel, color3f, matrix4d, point3f, float3, quatf, string}, caption=Example of a box with two flaps in USD format, captionpos=b, label=lst:example, frame=tb, aboveskip=0pt, framextopmargin=2pt, framexbottommargin=2pt]
		def Xform "world" () {
		def Xform "box" (
		prepend apiSchemas = ["PhysicsMassAPI", ...]) {
		matrix4d xformOp:transform = ( ... )
		token[] xformOpOrder = ["xformOp:transform"]
		float physics:mass = 2.79
		def Cube "geom_1" ( ... ) { ... }
		def PhysicsRevoluteJoint "box_flap_1_joint" {
		rel physics:body0 = </world/box>
		rel physics:body1 = </world/box_flap_1>
		}
		def PhysicsRevoluteJoint "box_flap_2_joint" { ... }
		}
		def Xform "box_flap_1" ( ... ) { ... }
		def Xform "box_flap_2" ( ... ) { ... }
		}
		\end{lstlisting}
	\end{lstfloat}
	
	In the remainder of this section, we will provide a more detailed definition of the characteristics of USD that are relevant for the translation method. 
	
	\subsection{Composed prims}
	
	We describe a scene graph as a set of prims which we denote as $\mathcal{P}$.
	Each prim $p \in \mathcal{P}$ is associated with a set of child prims.
	In Listing~\ref{lst:example}, child prims are used to represent different parts of a box, including the main part named \emph{box}, the geometry that form the box \emph{geom\_1} and the flaps named \emph{box\_flap\_1} and \emph{box\_flap\_2}.
	Each prim further has a specifier, a typed schema, an associated path, a set of API schemas and a set of properties.
	Accordingly, we say that each prim $p$ has the form $(h, t, n, P, M, E)$ where
	$h \in \mathcal{S}$ is the specifier,
	$t \in \mathcal{M}_\textrm{T}$ is the type schema,
	$n \in \mathcal{N}_\textrm{P}$ is the path,
	$P \subset \mathcal{P}$ are the children,
	$M \subset \mathcal{M}_\textrm{A}$ are the (applied) API schemas, and
	$E \subset \mathcal{E}$ are the (inherited) properties
	of the prim.
	Note that only a subset of properties from built-in schemas of the USD language is considered.
	Schemas used to describe shaders and rendering techniques are ignored.
	Here we focus on components of USD that are used in our example domain.
	Table~\ref{table} shows a selection of considered properties.
	
	\begin{table}
		\caption{A selection of properties considered in this work. 
		}\label{table}
		\centering
		\begin{tabular}{cccc}
			\toprule
			& \textbf{Schema} & \textbf{Property} & \textbf{Data type} \\
			
			\toprule
			\multirow{7}{*}{\rotatebox[origin=c]{90}{Typed schema}}
			& Xform 		& \textit{xformOpOrder} 				& \texttt{token[]} 		\\
			&      			& \textit{xformOp:transform} 			& \texttt{matrix4d} 	\\
			& Gprim 		& \textit{primvars:displayColor} 		& \texttt{color3f[]} 	\\
			&				& \textit{primvars:displayOpacity} 		& \texttt{float[]} 		\\
			& Cube			& \textit{size} 						& \texttt{double} 		\\
			& PhysicsJoint	& \textit{physics:body0} 				& \texttt{rel} 			\\
			&				& \textit{physics:body1} 				& \texttt{rel} 			\\
			
			\midrule
			\multirow{5}{*}{\rotatebox[origin=c]{90}{API schema}}
			& PhysicsMassAPI	& \textit{physics:mass} 				& \texttt{float} 		\\
			&				    & \textit{physics:centerOfMass} 		& \texttt{point3f} 		\\
			& RdfAPI			& \textit{rdf:namespace} 				& \texttt{string} 		\\
			&				    & \textit{rdf:conceptName} 				& \texttt{string} 		\\
			& SemanticTagAPI	& \textit{semanticTag:semanticLabel} 	& \texttt{rel} 			\\
			\bottomrule
		\end{tabular}
	\end{table}
	
	\paragraph{Specifier}
	Each prim is associated to a so called \emph{specifier} that is used to distinguish concrete and abstract prim declarations.
	We denote the set of prim specifiers as
	$\mathcal{S} = \{ \textit{def}, \textit{over}, \textit{cls} \}$.
	These specifiers are used for prims that describe concrete objects in the scene, to overwrite certain properties, and to define parent classes of prims respectively.
	In the listed example, only the \textit{def} specifier is used as each prim corresponds to a concrete object.
	
	\paragraph{Paths}
	USD uses paths to locate prims within a stage, starting with the root path \emph{"/"}. Each prim has a name string, and child prims are located using paths prefixed by their parent's path. 
	In the following, we denote the set of all possible prim names as $\mathcal{N}_\textrm{N}$, and assume that each prim name is prefixed by a slash.
	Hence, we can define the set of paths that can be constructed over these names as $\mathcal{N}_\textrm{P} = \mathcal{N}_\textrm{N}^*$.
	
	\paragraph{Schemas}
	In USD,  \emph{schemas} set out conventions and rules for structuring, organizing, and accessing data. Two types of schemas are distinguished:  \emph{typed schemas} and \emph{API schemas}. Typed schemas categorize the general types of prims, whereas API schemas specify certain attributes and relationships to describe object attributes. A prim can have only one typed schema but multiple API schemas. 
	We denote the set of typed schemas as $\mathcal{M}_\textrm{T}$.
	The set contains elements for all built-in and user-defined typed schemas and an additional constant $\notype \in \mathcal{M}_\textrm{T}$
	indicating that a prim has no typed schema.
	Similarly, we denote the set of API schemas as $\mathcal{M}_\textrm{A}$.
	The form of schemas is only relevant for resolving property values based on inheritance relations.
	
	
	\paragraph{Properties}
	Properties of prims define specific aspects of the associated object, such as its transformation or material.
	USD distinguishes between two types of properties: attributes and relationships.
	Accordingly, we see the set of properties $\mathcal{E}$ as the union of attributes and relations, i.e., 
	$\mathcal{E} = \mathcal{E}_\textrm{A} \cup \mathcal{E}_\textrm{R}$ where $\mathcal{E}_\textrm{A}$ is the set of attributes and $\mathcal{E}_\textrm{R}$ the set of relations.
	Attributes are seen as tuples of attribute name and value.
	Let $\mathcal{N}_\textrm{A}$ and $\mathcal{V}_\textrm{A}$ be the set of allowed attribute names and values respectively.
	Then each attribute $a \in \mathcal{E}_\textrm{A}$ can be written as a tuple $(n,v)$ where $n \in \mathcal{N}_\textrm{A}$ and $v \in \mathcal{V}_\textrm{A}$.
	In the example displayed in Listing~\ref{lst:example}, the prim named \emph{box} specifies four attributes.
	Their names often indicate in which schema they are defined, e.g. the \emph{xformOp} prefix indicates that attributes are derived from the \emph{Xform} schema.
	Prims can also have relationships with other prims.
	These are indicated by the keyword \emph{"rel"} in the USD format.
	The value of a relation is given by a set of USD paths, i.e. one relation can link to many paths.
	Hence, we can define the set of relations as $\mathcal{E}_\textrm{R} = \mathcal{N}_\textrm{R} \times \mathbb{P}(\mathcal{N}_\textrm{P})$ where $\mathcal{N}_\textrm{R}$ is the set of allowed relation names, and $\mathbb{P}(\mathcal{N}_\textrm{P})$ the powerset of paths.
	
	\subsection{Prim specifications}
	
	Prim specifications can be seen as prims in their uncomposed form, and as part of a layer without the context in which the layer is embedded.
	Composed prims are created through a composition method where many prim specifications may contribute to the composed description.
	Such a composed prim includes properties inherited from prim specifications and the schemas they refer to.
	Prim specifications, on the other hand, only include properties and schemas that are referred to in the scope of the specification itself, but they additionally specify \emph{composition arcs} which are used to compute the inheritance relation when prim compositions are created.
	Of the six composition arc types in USD, this paper focuses on the \emph{sublayer} and \emph{inherits} compositions. The sublayer composition integrates entire USD files into a main USD file, allowing the handling of complex scenes. 
	Meanwhile, 
	the inherits composition facilitates property inheritance among prims, aiding in creating variations and similar resources.
	
	The representation of prims and their specifications is very similar.
	For our purpose here, the notion of prims only needs to be extended to cover the inheritance relation.
	Accordingly, we
	denote the set of prim specifications in a layer as $\mathcal{P}_\textrm{S}$, and
	say that each prim specification $s \in \mathcal{P}_\textrm{S}$ has the form $(h, t, n, P, I, E)$ where
	$h \in \mathcal{S}$ is the specifier,
	$t \in \mathcal{M}_\textrm{T}$ is the type schema,
	$n \in \mathcal{N}_\textrm{P}$ is the path,
	$P \subset \mathcal{P}_\textrm{S}$ are the children,
	$I \in \mathbb{P}(\mathcal{M}_\textrm{A}) \times \mathbb{P}(\mathcal{N}_\textrm{´})$ is a tuple of API schemas and inheritance relations, and
	$E \subset \mathcal{E}$ are the properties
	of $s$.

	\section{Ontological Modeling of USD Environments}
	\label{sec:onto_modeling}
	
	The structure of KGs can be fixed through an ontological characterization of the vocabulary which is used to label nodes and edges.
	Here, we explore the use of an ontology language in the \emph{Description Logic} fragment of the \emph{Web Ontology Language} to support the translation process from USD format to a KG representation via logical axioms.
	The highest level of the ontology hierarchy we employ are externally defined ontologies that are widely applicable.
	The lower levels are aligned to the upper level, and thus must commit to the design decisions established at the upper level.
	
	\paragraph*{Upper-Level Ontology}
	For the most general definitions, we import the \emph{DOLCE+DnS Ultra-lite} (DUL) foundational framework~\cite{DOLCE2003}.
	It defines concepts such as \texttt{PhysicalObject} which is the class of all things that occupy space,
	and properties that can characterize objects further such as \textit{hasComponent} which is used to represent component hierarchies.
	DUL further defines a general model for object qualities around the notion of \texttt{Quality} that allow to form statements about individual aspects of entities.
	
	

	\subsection{The USD Ontology}
	The purpose of the USD ontology is to establish a graph model of the USD language, and to support translation from a USD to a KG scene representation.
	To this end, we define the USD vocabulary in terms of an ontology.
	In addition, a set of re-usable built-in schemas within the USD language is used to describe prims. In order to fasciliate translation, the ontological model attempts to replicate the structure of USD as closely as possible while in addition entailing semantics of USD entities.
	In the following, some central axioms of the ontology will be highlighted.
	However, due to space restrictions only a few will be listed in this paper.
	
	In its current version, the USD ontology consists of 35 concepts, 42 properties, and 213 logical axioms.
	The ontology characterizes the composed view on prims.
	The layered view is not considered as it is mostly relevant for the authoring of environment models.
	The ontology further covers a set of built-ins of the USD language.
	These are 3 API schemas, and 11 concepts representing different typed schemas
	covering the requirements of the usecase considered in this work.
	However, many other built-in schemas can be characterized analogously.
	
	\subsubsection{Representation of prims}
	The basic notion in the USD ontology is the \texttt{Prim} concept.
	It represents concretely defined prims (i.e., using the \emph{def} keyword).
	Prims with other specifiers (i.e., \emph{over} and \emph{class}) that only contribute to the uncomposed view are not considered in this work.
	In terms of the toplevel ontology, we define that
	$\texttt{Prim} \sqsubseteq \texttt{Object}$, where the notion of \texttt{Object} also includes objects without a proper space region.
	Parthood of prims is represented using the transitive and reflexive \textit{hasPart} relationship of the toplevel ontology.
	
	Each prim has a set of data attributes and relationships.
	Relations trivially map to edges between prims in the KG.
	Data attributes are used in the USD language to quantify characteristics of prims.
	The attribute \textit{xformOp:scale}, for example, is used to quantify how a prim is scaled along three dimensions.
	It thus contributes to the shape of the prim.
	Following the toplevel ontology, we represent such characteristics using the \texttt{Quality} concept, e.g., $\texttt{Shape} \sqsubseteq \texttt{Quality}$ represents the quality of having a shape.
	Each quality concept further defines a property used to connect a prim to its quality:
	\begin{eqnarray}
	\label{eq:shape0} \textit{hasShape} &\sqsubseteq& \textit{dul:hasQuality} \\
	\label{eq:shape1} \textit{hasShape}(x,y) &\rightarrow& \texttt{PO}(x) \wedge \texttt{Shape}(y) \\
	\label{eq:shape2} \texttt{Prim} &\sqsubseteq& (\leq 1) \textit{dul:hasQuality}.\texttt{Shape}
	\end{eqnarray}
	Axiom~(\ref{eq:shape1}) defines range and domain of the property \textit{hasShape}.
	Namely it asserts that having a shape implies that a prim must be of type \texttt{dul:PhysicalObject} (short \texttt{PO}).
	Axiom~(\ref{eq:shape2}) asserts that a prim may have at max one shape.
	
	We further define a set of data types of the USD language in the ontology such that they can be used to represent typed literals in the KG.
	This includes data types such as \textit{float3} and \textit{float4} that represent numeric vectors.
	The datatypes and quality concepts are then used to define built-in attributes of the USD language.
	For example,
	\textit{xformOp:scale} is defined as an attribute of the \texttt{Shape} quality of a prim where the value of the attribute has the data type \texttt{float3}.

	\subsubsection{Representation of schemas}
	Most of the built-in schemas of the USD language are used to describe prims, but a few also exist that describe properties.
	We characterize schemas ontologically using the notion of \texttt{Description}. 
	Each schema entails that a certain set of properties is used for describing some phenomena such as properties used to quantify the mass of an object.
	As such schemas describe objects where certain qualities of the object are quantified in a specific way.
	They may further imply type constraints for objects they describe.
	
	The API schema and typed schema (also called IsA schema) in the USD language is represented via the concepts \texttt{APISchema} and \texttt{TypedSchema} respectively.
	Both follow the same pattern how they are characterized:
	they may restrict the type of prims described by them, and further assert a particular quantification for a quality of the prim.
	For example, we define the \texttt{Xformable} schema as follows:
	\begin{eqnarray}
	\label{eq:xform0} \texttt{Xformable} &\sqsubseteq& \texttt{TypedSchema} \\
	\label{eq:xform1} \texttt{Xformable} &\sqsubseteq& \forall \textit{isSchemaOf}.\texttt{PhysicalObject} \\
	\label{eq:xform2} \texttt{Xformable} &\sqsubseteq& \forall \textit{isSchemaOf}.\texttt{WithXform} \\
	\label{eq:xform3} \texttt{WithXform} &\equiv& \exists \textit{hasShape}.
	(\exists \textit{xformOp:scale} \sqcap \dots)
	\end{eqnarray}
	Hence, only prims with type \texttt{PhysicalObject} can have this schema (\ref{eq:xform1}), and it establishes the use of data properties for the quantification of the shape (\ref{eq:xform2}, \ref{eq:xform3}).
	Further note that \textit{isSchemaOf} is aligned to the toplevel via its parent property \textit{describes}.
	
	Qualities of prims can also be classified via schemas.
	For instance, $\texttt{CubeShape} \sqsubseteq \texttt{Shape}$ represents the quality of having a cube-like shape.
	We define \texttt{CubeSchema} as a schema that must have a \texttt{CubeShape}.
	Hence, any shape of a \texttt{CubeSchema} can be classified as a \texttt{CubeShape} which enables the classifcation via reasoning.
	
	
	\subsubsection{Connectedness of Prims}
	Prims are thought to be directly connected if there exists a joint that connects them.
	Joints are represented as prims that are described by a \texttt{PhysicsJointSchema}.
	The schema establishes the use of the relations $\textit{physics:body0} \sqsubseteq \textit{hasConnection}$ and $\textit{physics:body1} \sqsubseteq \textit{hasConnection}$ to link joint with prims:
	\begin{eqnarray}
	\label{eq:conn0} \textit{hasConnection}(x,y) &\rightarrow& \textit{hasConnection}(y,x) \\
	\label{eq:conn1} \textit{physics:body0} &\sqsubseteq& \textit{hasConnection} \\
	\label{eq:conn2} \textit{physics:body1} &\sqsubseteq& \textit{hasConnection}
	\end{eqnarray}
	Hence, the (direct) connection between a joint and the two prims it controls is bidirectional (\ref{eq:conn0}).
	
	However, prims of a joint are also connected with each other via the joint.
	This indirect connectedness can be inferred through the introduction of $\textit{HC}^\textrm{T}$ which is a transitive variant of the property:
	\begin{eqnarray}
	\label{eq:conn3} \textit{hasConnection} &\sqsubseteq& \textit{HC}^\textrm{T} \\
	\label{eq:conn4} \textit{HC}^\textrm{T}(x,y) \wedge \textit{HC}^\textrm{T}(y,z) &\rightarrow& \textit{HC}^\textrm{T}(x,z)
	\end{eqnarray}
	
	\subsection{Logical Theories for Robot Decision Making}
	\label{subsec:boxunpacking}
	
	Formal ontologies enable the use of logical reasoning as they define a theory around notions they consider.
	The USD ontology encodes such a theory which is designed for reasoning about the mapping between USD and KG representations.
	Additional theories can be linked to the USD ontology to support robot decision making.
	As a proof of concept, in the scope of this work, we define an object type \texttt{Box} with two flaps of type \texttt{Flap} can be classified as either \emph{opened} or \emph{closed} based on a fixed threshold of the joint value connecting the box with the flaps.
	Accordingly, we assert that $\textit{hasQuality}.(\exists \textit{hasJointValue}.(\geq 0.1))$ is a sufficient condition for a joint to be classified as \emph{opened}, and that a \texttt{Box} can be classified as \emph{opened} whenever its joints are opened.
	Axioms for being closed are defined analogously.
	
	\section{Knowledge Graph Construction}
	\label{sec:KG_construction}
	\newcommand\mycommfont[1]{\scriptsize\ttfamily\textcolor{blue}{#1}}
	\SetCommentSty{mycommfont}
	
	
	The first step in our proposed method for generating a KG representation of a USD scene is to establish a USD layer that includes a collection of class prims representing the TBox ontology.
	Another USD layer representing the scene graph imports the TBox layer, and uses a custom API for tagging prims with ontological concepts.
	The semantically tagged prims are then translated into the KG.
	
	
	
	\subsection{USD sublayer for semantic tagging}
	\label{subsec:USD_sublayer}
	
	In order to link prims into the KG, we introduce the \emph{SemanticTagAPI} schema.
	It views semantic tags as relationships in the USD framework, and uses the \emph{semanticLabel} property to link prims to class prims corresponding to ontological concepts.
	The class prims utilize the \emph{RdfAPI} schema to represent identifier of entities in the KG using the properties \emph{rdf:namespace} and \emph{rdf:conceptName}.

	The USD layer for semantic tagging can then easily be generated given an ontology.
	A formalization of this procedure is omitted here due to space restrictions.
	In principle, it
	converts a set of concept names 
	into a USD layer 
	consisting prim specifications.
	For each concept in the ontology, a class prim is created according to the class hierarchy in the ontology, the concept name, and its associated namespace.
	The main USD layer that requires semantic tagging may then make use of class prims from the generated sublayer to classify prims.

	\subsection{USD to KG translation}
	\label{subsec:usd_to_KG}
	
	\begin{algorithm}[h]
		\SetKwFunction{isOddNumber}{isOddNumber}
		\SetKwInOut{KwIn}{Input}
		\SetKwInOut{KwOut}{Output}
		\SetKwFunction{KwFn}{Fn}
		\SetNoFillComment
		
		\KwIn{A set of prims $\mathcal{P}$ forming a USD stage.}
		\KwOut{An ABox that describes the prims in $\mathcal{P}$.}
		
		$\abox \leftarrow \emptyset$ \tcp{an empty ABox}
		
		\For{$(\textsf{def}, t, n, P, M, E) \in \mathcal{P} \wedge i = \textsf{iri}(n)$}
		{
			\If(\tcp*[h]{assert $t$ as type of $i$}){$t \neq \notype \wedge c = \textsf{iri}_\textrm{S}(t)$}
			{
				$\abox \leftarrow \abox \cup \{
				(\exists \textit{hasTypedSchema}.c)(i))
				\}$
			}
			
			\For(\tcp*[h]{assert parts of $i$}){$(\textsf{def}, \_, n_c, \_, \_, \_) \in P$}
			{
				$\abox \leftarrow \abox \cup \{ \textit{hasPart}(i, j) \}$ where $j = \textsf{iri}(n_\textrm{c})$
			}
			
			\For(\tcp*[h]{assert API schema $m$ of $i$}){$m \in M$}
			{
				$\abox \leftarrow \abox \cup \{
				(\exists \textit{hasAPI}.c)(i))
				\}$, $c = \textsf{iri}_\textrm{S}(m)$
			}
			
			\For{$(n_\textrm{E}, v_\textrm{E}) \in E \wedge i_E = \textsf{iri}_P(n_\textrm{E})$}
			{
				\uIf{$n_\textrm{E} = \textit{semanticTag:semanticLabel}$}
				{
					\tcp{assert semantic tags as types of $i$}
					
					\For{$n_\textrm{r} \in v_\textrm{E} \wedge (\textsf{cls}, \_, n_\textrm{r}, \_, \_, F) \in \mathcal{P}$}
					{
						$i_\textrm{t} \leftarrow b \oplus g$, where $(\textit{rdf:ns}, b) \in F, (\textit{rdf:name}, g) \in F$
						
						$\abox \leftarrow \abox \cup \{ i_t(i) \}$
					}
				}
				\uElseIf(\tcp*[h]{assert relation of $i$}){$n_\textrm{E} \in \mathcal{N}_R$}
				{
					\For(\tcp*[h]{to a prim at path $n_r$}){$n_r \in v_\textrm{E}$}
					{
						$\abox \leftarrow \abox \cup \{ i_\textrm{E}(i, j) \}$ where $j = \textsf{iri}(n_\textrm{r})$
					}
				}
				\Else
				{
					
					$\abox \leftarrow \abox \cup \{
					\textit{hasQuality}(i, i_\textrm{Q}),
					i_E(i_\textrm{Q}, v_\textrm{E})
					\}$ where $i_\textrm{Q} = \textsf{iri}(\texttt{Quality})$
				}
			}
		}
		\Return $\abox$
		
		\caption{USD to KG translation}
		\label{algo:usd_to_KG}
	\end{algorithm}
	
	This subsection details the transformation of a USD layer into a KG.
	Algorithm \ref{algo:usd_to_KG} formalizes this process for a set of prims $\mathcal{P}$.
	To this end, an ABox ontology $\abox$ is genertated that contains facts representing the prims in $\mathcal{P}$.
	For simplicity, we assume that $\mathcal{P}$ also includes class prims (not only defined prims).
	The ABox is initially empty, and successively populated during an iteration of prims in $\mathcal{P}$.
	In the following, several assertions about each prim in $\mathcal{P}$ are added to the ABox $\abox$.
	These are statements involving the prim identifier $i$ as a subject which identifies the prim at path $n$ in the KG.
	The path $n$ is mapped via the function $\textsf{iri}$ to an identifier of the prim in the KG.
	The resulting ABox $\abox$ contains named individuals representing prims in $\mathcal{P}$ and their qualities.

	\section{Evaluation}
	
	This section demonstrates how USD supports robotics agents to perform human-scale tasks. The evaluation utilizes the robots simulation from MuJoCo \cite{todorov2012mujoco}, which is controlled by Giskard \cite{giskard}, a constraint-based motion planning using ROS communication and visualized by Unreal Engine.  It also highlights USD's role in parsing scene descriptions between MuJoCo, ROS, Unreal Engine, and translating into the KG. 
	
	\subsection{Universal Scene Description Parser}
	
\begin{figure}[t]
		\centering
		\includegraphics[width=\columnwidth]{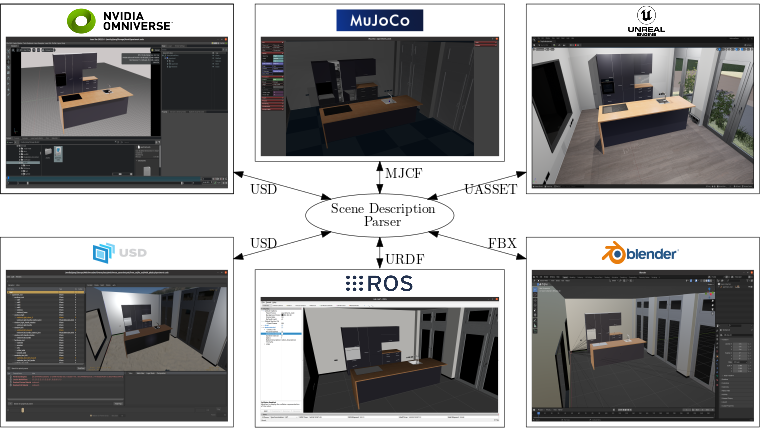}
		\caption{Usage of USD as the translation medium for scene descriptions across various formats}
		\label{fig:MultiverseParser}
	\end{figure}

	Utilizing USD for scene description parsing offers consistency and minimal data loss. Analogous to the KG translation process, any new attributes specified in the scene description format, which USD doesn't originally recognize, can be incorporated into USD by introducing a new schema API. In this work, we only process commonly shared data, including transformations, geometry, and inertial properties of bodies and their associated joints. 
	In essence, given a scene description in USD or any other format, this parser optimizes and translates it into USD, streamlining its conversion into KG or another format, as depicted in Figure \ref{fig:MultiverseParser}.
	
	\subsection{Giskard}
	Giskard \cite{giskard} is a constraint-based motion planning and control framework.
	In contrast to many alternatives, it models not only the robot, but also the world kinematics\cite{rofer2022kineverse}.
	This simplifies the expression of environment manipulation motions, by talking in terms of how the environment should change.
	Additionally, the state of Giskard's world model can be synchronized by reading the joint state from the simulation environment or estimating it in reality.
	
	\subsection{Task execution with USD}
	
\begin{figure}[h]
		\centering
		\includegraphics[width=\columnwidth]{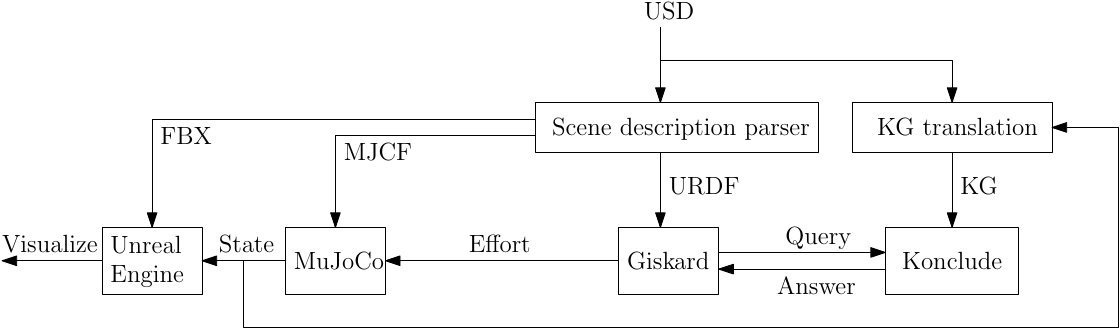}
		\caption{Task execution pipeline from knowledge to action}
		\label{fig:Demo}
	\end{figure}

	Robot's comprehensive awareness of its environment is important for its performance in human-scale tasks. For tasks like tidying objects, robots might rely on online information or user cues, like "milk should be stored in the fridge". Using USD and the built KG, robots execute tasks with a well-formed belief state, increasing their adaptability in human-centered tasks. In the context of our experiment, the robot is directed to open a box, identify its contents, and neatly place them in a proper location.  
	The task execution pipeline \footnote{\url{https://github.com/Multiverse-Framework/Multiverse-Docker/tree/ICRA-2024}} is depicted in Figure \ref{fig:Demo}.
	
	During the initialization phase, the scene graph of the apartment, consisting of 2247 prims (784 of type \emph{Xform}, 658 of type \emph{Gprim}, 45 of type \emph{Joint}), is translated into the KG as an OWL file with 1487 prims, excluding 760 prims of type \emph{Material}, \emph{Shader}, etc. Subsequently, the KG with 6827 nodes and 24116 edges was subjected to a reasoning process to evaluate its consistency. As tasks are executed, the simulation updates the state to the KG, which is accessed by Konclude \cite{Konclude} for queries. Using Konclude's \emph{realization} operation, one can determine an individual's class, helping label items in a scene — for instance, determining if a flap is open or if an item is perishable. Another query identifies suitable storage locations for items, based on logic like "every spoon can be stored in any drawer". While OWL-DL and Konclude have limitations in handling this, we use an ontology of item uses \cite{Narrative}. This combines OWL-DL's object taxonomy with rules about how object classes can interact for tasks.
	
\begin{figure}[h]
		\centering
		\includegraphics[width=\columnwidth]{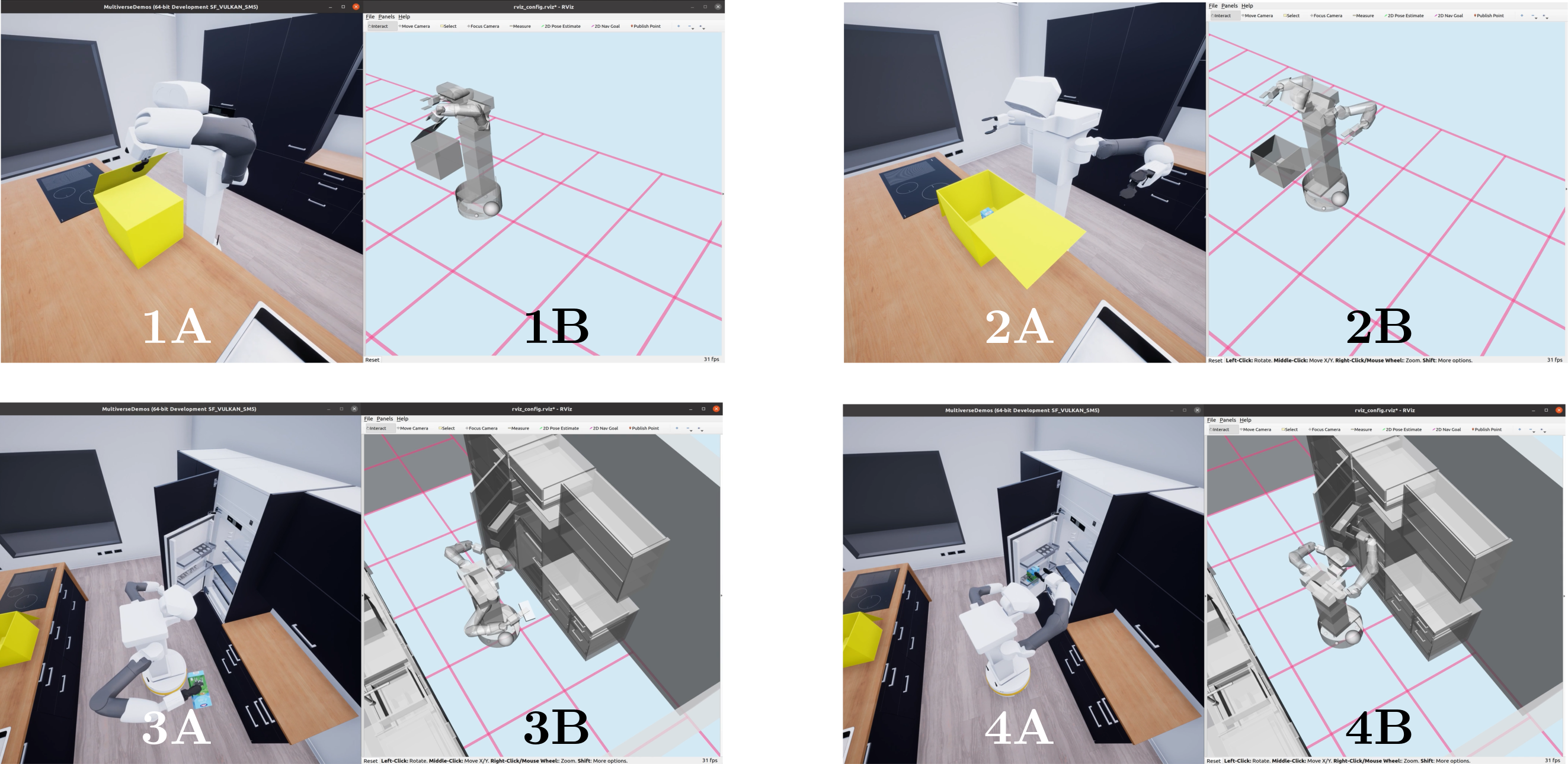}
		\caption{Box unpacking scenario (Left: Unreal Engine, Right: RViz)}
		\label{fig:Demo2}
	\end{figure}

	Giskard controls the robot using simulation data and constraints. Task decisions are derived from Konclude's reasoning, which informs Giskard about the box's status and object storage location. As the result, the robot successfully executes the task utilizing the scene graph-grounded KG. Figure \ref{fig:Demo2} illustrates the task execution during each phase when Konclude updates Giskard's belief state. 
	
	On the left, the visualization in Unreal Engine is depicted, whereas the right side represents the simulated robot's belief state of Giskard in RViz. In the first phase, the box's state is queried, disregarding the object inside it and the apartment. In the second phase, after the object \emph{milk} is identified, the object's storage container \emph{fridge} is determined. In phases three and four, the robot picks the \emph{milk}, opens the \emph{fridge}, places it in the \emph{tray}, and then closes the \emph{fridge}.

	\section{Conclusion}
	This research presents a method for translating USD scene graphs into KGs, addressing three crucial aspects of the process. Firstly, the formalization and ontological modeling of the USD scene graph enables its representation in the KG and allows for the inference of additional knowledge. Secondly, a sublayer USD is generated, which represents the TBox ontology and enables semantic tagging using custom API schemas. The main USD layer is then linked to the knowledge base using these API schemas, and covers all individuals relevant for reasoning. Finally, the USD layer is converted into a KG, augmented with dynamical data from a physics simulator. 
	By linking the USD into the KG, background knowledge for certain robotic decision-making tasks, as demonstrated in a box unpacking scenario.
	
	\section*{Acknowledgment}
	
	\thanks{The research reported in this paper has been supported by the trilateral Project \#442588247 ''ai4hri – artificial intelligene for human-robot interaction'' which is partly funded by the German DFG.}
	
	%
	%
	%
	%
	
	\bibliographystyle{amsplain}
	\bibliography{references.bib}
	
\end{document}